\documentclass[11pt]{article}
\usepackage[margin=1in]{geometry}
\usepackage{times}
\usepackage{microtype}
\usepackage[numbers,square,sort&compress]{natbib}
\usepackage{graphicx}
\usepackage{booktabs}
\usepackage{amsmath,amssymb}
\usepackage{multirow}
\usepackage{xcolor}
\usepackage{url}
\usepackage[hidelinks]{hyperref}
\usepackage{caption}
\usepackage{subcaption}
\usepackage{tikz}
\usetikzlibrary{arrows.meta,positioning,fit,backgrounds,calc}
\title{FANVIDv2: Evaluating Video Super-Resolution by Face and\\
Licence-Plate Recognition Under Compound Degradation}

\author{Kavitha Viswanathan\thanks{Corresponding author.} \and
Vrinda Goel \and Shlesh Gholap \and Devayan Ghosh \and
Madhav Gupta \and Dhruvi Ganatra \and Sanket Potdar \and Amit Sethi\\
\small Department of Electrical Engineering, Indian Institute of Technology Bombay, Mumbai 400076, India}
\date{}

\begin{document}
\maketitle

\begin{abstract}
Video super-resolution (VSR) is normally judged by PSNR and SSIM on clips that
were downsampled bicubically, although in surveillance its purpose is to make
faces and licence plates \emph{recognisable}. We present FANVIDv2, a benchmark
that scores VSR by what a recognition pipeline can do with its output.
FANVIDv2 provides $320\times180$ low-resolution (LR) clips with
high-resolution (HR) references for 48 public figures (with one HR gallery
image each) and 375 licence-plate clips covering 360 distinct plate strings.
LR clips are generated with a randomised compound degradation (blur, resize
jitter, sensor noise, JPEG compression, final downsampling) rather than
bicubic downsampling alone. Two metrics score recognition \emph{inside}
detections: FaceRecBox rewards a face only if it is localised and correctly
identified, and TextRecBox scores plate transcriptions by normalised edit
distance weighted by localisation quality. With a 2.3\,M-parameter VSR baseline
(RCDM), FaceRecBox rises from 0.6864 to 0.7222, identity accuracy on matched
faces from 84.35\% to 86.93\%, and TextRecBox from 0.3088 to 0.3667; a
residual-map gated variant (RCDM-RMGF) reaches 0.3801 on plates. We describe
the degradation model, the baseline architectures and the scorers in detail,
and release annotations, metadata, download and degradation scripts and
evaluation code.
\end{abstract}

\section{Introduction}
Surveillance video is often too coarse for its purpose: a face that occupies a
few dozen pixels, or a plate whose characters blur together, cannot be
recognised in any single frame \citep{bbc2025}. Video super-resolution (VSR)
could help, because the missing detail is partly present in neighbouring
frames. Current VSR benchmarks do not measure this. REDS \citep{nah2019reds},
Vimeo-90K \citep{xue2019toflow} and Vid4 \citep{liu2014bayesian} generate
low-resolution (LR) inputs by bicubic downsampling of clean video and score
outputs with PSNR and SSIM. Real cameras add blur, noise and compression, and a
reconstruction can gain PSNR while altering exactly the high-frequency detail
that decides whether a character is a B or an 8 \citep{blau2018perception}.

FANVIDv2 evaluates VSR by recognition. It extends FANVID
\citep{viswanathan2025fanvidv1}, which introduced face and plate clips with
bicubic LR inputs, in three ways: LR inputs are generated with a randomised
compound degradation (Section~\ref{sec:degr}); both tasks are scored with
metrics that couple detection and recognition (Section~\ref{sec:metrics});
and the benchmark is accompanied by a reference pipeline, a lightweight VSR
baseline family and open scorers (Sections~\ref{sec:pipeline}--\ref{sec:results}).
Our contributions are:
\begin{enumerate}
\item A benchmark of face clips for 48 identities, each with an HR gallery
image, and 375 plate clips with 360 distinct plate strings
(Section~\ref{sec:data}).
\item A randomised compound degradation model, its released implementation,
and the second-order variant used for training (Section~\ref{sec:degr}).
\item FaceRecBox and TextRecBox and their released scorers (Section~\ref{sec:metrics}).
\item A documented family of memory-, wavelet- and deformable-convolution-based
VSR baselines with baseline results for LR input, RCDM and RCDM-RMGF that
report recognition, localisation and OCR counts, with rates normalised for
differing numbers of evaluated frames (Sections~\ref{sec:pipeline}--\ref{sec:results}).
\end{enumerate}

\section{Related work}
\textbf{VSR benchmarks.} REDS, Vimeo-90K and Vid4 use bicubic or
blur-and-downsample degradations and fidelity metrics. Real-ESRGAN
\citep{wang2021realesrgan} and BSRGAN \citep{zhang2021bsrgan} introduced
randomised, higher-order degradation models for blind SR; we use models of this
kind both to generate test inputs and to train the baseline.
Large VSR models such as BasicVSR++ \citep{chan2022basicvsrpp}, VRT
\citep{liang2024vrt} and RVRT \citep{liang2022rvrt} are evaluated almost
exclusively with fidelity metrics.

\textbf{Surveillance recognition datasets.} Person re-identification
benchmarks such as iLIDS-VID \citep{wang2014ilidsvid} and MARS \citep{mars}
provide multi-frame tracklets, but match whole bodies at native camera
resolution without SR. SCface \citep{scface} and QMUL-SurvFace
\citep{cheng2018survface} contain low-resolution surveillance faces but are
single-frame and do not evaluate SR. Plate datasets such as UFPR-ALPR
\citep{ufpr} and CCPD \citep{ccpd} provide plates at resolutions sufficient
for direct OCR.

\textbf{Task-aware evaluation of SR.} Restoration quality and recognition
utility can diverge \citep{blau2018perception}. FANVIDv2 provides a common
testbed in which this divergence can be measured for two recognition tasks with
different requirements: faces need global structural coherence, plates need
character-level precision.

\section{The FANVIDv2 dataset}\label{sec:data}
\subsection{Sources and content}
All clips are cut from publicly available YouTube videos; only URLs,
timestamps and annotations are distributed, and frames are regenerated with the
released download scripts (Section~\ref{sec:release}). HR frames are
$1280\times720$; LR frames are $320\times180$ (scale $\times4$).

\textbf{Faces.} Clips show 48 public figures of diverse nationalities, genders
and ethnicities in public appearances. Each
identity has one HR gallery image taken from a different source (press
photography), and the face gallery is released as a name--image table.
Training and test identities are disjoint.

\begin{figure}[t]
\centering
\includegraphics[width=0.9\columnwidth]{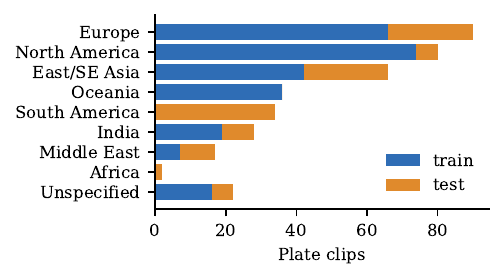}
\caption{Plate clips by region and split, computed from the released metadata.
Twenty-two clips have no region label.}\label{fig:regions}
\end{figure}

\textbf{Plates.} The plate subset contains 375 clips from dash-camera and
street videos, with 360 distinct plate strings. Plates are Latin alphanumeric
and come from more than 25 cities on five continents. The split is by clip
(260 training and 115 test clips); two plate strings occur in both splits.
Fig.~\ref{fig:regions} shows the geographic composition. The split
deliberately places some regions only in the test set (all Brazilian clips) and
others only in training (all Oceanian clips), so the test set contains plate
formats not seen in training.

\begin{table}[t]
\caption{FANVIDv2 composition (frame counts are LR frames of annotated clips).}
\label{tab:composition}
\centering\small
\begin{tabular}{@{}lccccc@{}}
\toprule
Subset & IDs / strings & Clips & Train frames & Test frames & Eval.\ GT boxes\\
\midrule
Faces (public figures) & 48 & 828 & 101,870 & 48,500 & 139,093\\
Licence plates & 360 & 375 & 47,806 & 18,292 & 34,996$^\dagger$\\
\bottomrule
\end{tabular}

\vspace{2pt}
{\footnotesize $^\dagger$ Ground-truth boxes are counted on the frames common to the HR annotations and the system output: 34,996 for SR output and 31,739 for LR output (Section~\ref{sec:results}).}
\end{table}

\subsection{Annotation}
Faces are detected on HR frames with RetinaFace \citep{retinaface}, embedded
with ArcFace \citep{arcface}, matched to the gallery, and every retained box
and identity is checked and corrected manually; a second annotator
cross-checked identity labels (inter-annotator agreement $\kappa=0.91$). Plates are annotated
manually on HR key frames, propagated through the clip with SAM\,2
\citep{sam2}, and reviewed frame by frame under an IoU threshold of 0.7;
plate strings are transcribed manually and normalised to upper-case
alphanumerics. Faces use an IoU threshold of 0.5. Clips also contain distractor faces
and plates, and predictions on them count as false positives for faces.

\section{Compound degradation}\label{sec:degr}
Bicubic downsampling produces LR frames that are unrealistically clean, and
models trained on them collapse on real footage. FANVIDv2 therefore uses two
related degradation models (Fig.~\ref{fig:degr}), one to \emph{generate the
benchmark's LR clips} and one, richer, to \emph{train} VSR models.

\begin{figure}[t]
\centering
\resizebox{\textwidth}{!}{%
\begin{tikzpicture}[
  node distance=11mm,
  blk/.style={draw,rounded corners=2pt,minimum height=9mm,minimum width=15mm,align=center,font=\scriptsize,fill=blue!6},
  io/.style={draw,minimum height=9mm,minimum width=14mm,align=center,font=\scriptsize,fill=gray!15},
  arr/.style={-{Stealth[length=2mm]},thick},
  lab/.style={font=\tiny,align=center,text=black!70,text width=23mm}]
\node[font=\small\bfseries,anchor=west] (ta) at (0,1.5) {(a) Benchmark LR generator (released script; sampled independently per frame)};
\node[io] (hr) at (0,0) {HR frame\\$1280{\times}720$};
\node[blk,right=of hr] (b1) {Blur\\$p{=}0.5$};
\node[blk,right=of b1] (b2) {Resize\\jitter};
\node[blk,right=of b2] (b3) {Noise\\$p{=}0.5$};
\node[blk,right=of b3] (b4) {JPEG};
\node[blk,right=of b4] (b5) {Bicubic\\$\downarrow4$};
\node[io,right=of b5] (lr) {LR frame\\$320{\times}180$};
\foreach \a/\b in {hr/b1,b1/b2,b2/b3,b3/b4,b4/b5,b5/lr} \draw[arr] (\a)--(\b);
\node[lab,below=1mm of b1] {Gaussian ($p{=}.7$,\\$k\in\{3,5,7\}$, $\sigma\in[.2,2]$)\\or 1-D motion box};
\node[lab,below=1mm of b2] {factor $s\in[.5,1.5]$,\\random interp., resized\\back to original size};
\node[lab,below=1mm of b3] {Gaussian ($p{=}.7$,\\$\sigma\in[1,15]$) or\\Poisson};
\node[lab,below=1mm of b4] {quality\\$q\in\{30..95\}$};
\node[font=\small\bfseries,anchor=west] (tb) at (0,-3.1) {(b) Second-order training degradation (Real-ESRGAN style; drawn once per group of frames)};
\node[io] (hr2) at (0,-4.6) {HR GOF\\$I^{\mathrm{HR}}$};
\node[blk,right=6mm of hr2,fill=orange!10] (c1) {Blur\\$\kappa_1$};
\node[blk,right=of c1,fill=orange!10] (c2) {Resample\\$\downarrow s_1$};
\node[blk,right=of c2,fill=orange!10] (c3) {Noise\\$\eta_{\sigma_1}$};
\node[blk,right=of c3,fill=green!8] (c4) {Blur\\$\kappa_2$};
\node[blk,right=of c4,fill=green!8] (c5) {Resample\\$\downarrow s_2$};
\node[blk,right=of c5,fill=green!8] (c6) {Noise\\$\eta_{\sigma_2}$};
\node[blk,right=of c6,fill=green!8] (c7) {JPEG\\$q$};
\node[io,right=of c7] (lr2) {LR GOF\\$320{\times}180$};
\foreach \a/\b in {hr2/c1,c1/c2,c2/c3,c3/c4,c4/c5,c5/c6,c6/c7,c7/lr2} \draw[arr] (\a)--(\b);
\node[lab,below=1mm of c1] {iso./aniso./gen.\\Gaussian, plateau};
\node[lab,below=1mm of c4] {same families\\as stage 1};
\node[lab,below=1mm of c2] {nearest, bilinear,\\bicubic, area};
\node[lab,below=1mm of c3] {Gaussian/Poisson/\\colour mixture};
\node[lab,below=1mm of c7] {$q\sim\mathcal U[30,95]$};
\node[lab,below=1mm of c5] {as stage 1};
\begin{scope}[on background layer]
 \node[draw,dashed,rounded corners,fit=(c1)(c3),inner sep=2mm,label={[font=\tiny]above:stage 1}] {};
 \node[draw,dashed,rounded corners,fit=(c4)(c7),inner sep=2mm,label={[font=\tiny]above:stage 2}] {};
\end{scope}
\end{tikzpicture}}
\caption{Degradation models. (a) The LR clips of the benchmark are produced by a
single randomised pass, followed by a bicubic $\times4$ downsample (parameter
ranges as in the released script). (b) The second-order model used to
train the VSR baselines applies two randomised stages and draws its
parameters once per group of frames (GOF), preserving temporal correlation that a
recurrent memory relies on.}\label{fig:degr}
\end{figure}

\subsection{Benchmark LR generator}
Each HR frame is degraded by one randomised pass (Fig.~\ref{fig:degr}a), in
this order. \emph{Blur}: with probability 0.5, a Gaussian blur (probability
0.7; kernel size $k\in\{3,5,7\}$, $\sigma\sim\mathcal U[0.2,2.0]$) or a
horizontal box (motion) kernel of size $k$. \emph{Resize jitter}: the frame is
resized by a factor drawn from $\mathcal U[0.5,1.5]$ with an interpolation
method drawn from \{linear, cubic, area, Lanczos\} and resized back to its
original size. \emph{Noise}: with probability 0.5, additive Gaussian noise
(probability 0.7, $\sigma\sim\mathcal U[1,15]$ on the 0--255 scale) or Poisson
noise. \emph{JPEG}: compression at an integer quality drawn from $[30,95]$.
Finally the frame is downsampled bicubically by $\times4$ and resized to the
exact LR size. Because every parameter is drawn at random, the test set spans a
continuum from nearly clean to heavily corrupted frames rather than one
operating point. The implementation is in
\texttt{code/dataset\_script\_celebs.py} (faces) and
\texttt{code/download\_script\_lp.py} (plates).

\subsection{Second-order training degradation}
Following Real-ESRGAN \citep{wang2021realesrgan}, VSR training uses two
randomised stages,
\begin{align}
I^{(1)} &= \big(I^{\mathrm{HR}}\circledast\kappa_1\big)\!\downarrow_{s_1}\!+\,\eta_{\sigma_1},\\
I^{(2)} &= \mathrm{JPEG}_q\Big(\big(I^{(1)}\circledast\kappa_2\big)\!\downarrow_{s_2}\!+\,\eta_{\sigma_2}\Big),
\end{align}
with kernels $\kappa_{1,2}$ drawn from isotropic and anisotropic Gaussian,
generalised Gaussian and plateau families, scales $s_{1,2}$ realised by a mixed
\{nearest, bilinear, bicubic, area\} resampler, noise $\eta$ from a
Gaussian/Poisson/colour mixture, and $q\sim\mathcal U[30,95]$. For video, the
parameters are drawn \emph{once per GOF} rather than per frame, so that
temporal correlations within the input window are preserved for a recurrent
model.

\section{Tasks and metrics}\label{sec:metrics}
Let ground-truth boxes be indexed by $i$ and predicted boxes by $j$, with
intersection-over-union $I_{ij}$. Matching is greedy and one-to-one in
decreasing IoU.

\textbf{Face task (T1).} Given an LR clip and the HR gallery, a system detects
faces in each frame and assigns an identity or rejects the face. Let
$M_{ij}=\mathbf 1[I_{ij}\ge0.5]$ be the (one-to-one) matching indicator,
$F_{ij}\in\{0,1\}$ indicate a correct identity, and $N_i$ and $P_j$ indicate
unmatched ground-truth and predicted boxes. Then
\begin{equation}
\mathrm{FaceRecBox}=\frac{\sum_{ij}M_{ij}F_{ij}}{\sum_{ij}M_{ij}+\sum_iN_i+\sum_jP_j},
\end{equation}
i.e.\ correct identities divided by ground-truth boxes plus false positives. A
detection earns credit only if it is in the right place \emph{and} names the
right person; missed faces and spurious detections both count against it. We
also report precision, recall, F1, mean IoU of matches and identity accuracy
on matched faces.

{\sloppy
{\sloppy\textbf{Plate task (T2).} Given an LR clip, a system localises each plate and
transcribes it, with no list of valid plates. For a prediction $\hat s$ and
ground truth $s$ (both normalised to upper-case alphanumerics) the text score
is
\begin{equation}
T(\hat s,s)=\max\!\Big(0,\,1-\frac{\mathrm{Lev}(\hat s,s)}{|s|}\Big),
\end{equation}
where Lev is the Levenshtein distance. We use two scorers, both released.
The \emph{hard-IoU} scorer (\texttt{fanvid\_metrics\_arxiv.py}) follows FANVID
v1: a prediction counts only if $I_{ij}\ge0.5$, and false positives are excluded
from the denominator because plate strings are open-vocabulary. The
\emph{aligned} scorer (\texttt{Fanvid\_metrics\_aligned.py}), used for all
plate results in this paper, additionally (i) restricts evaluation to frames
present in both the HR annotations and the system output, which removes
frame-timing mismatches between HR annotations and LR/SR predictions, and
(ii) weights each matched prediction by its localisation quality: the IoU is
mapped linearly to a weight $w_{\mathrm{IoU}}\in[0,1]$ (zero at $I{=}0.2$, one at $I{=}1$) and
the box score is $0.5\,w_{\mathrm{IoU}}+0.5\,T$. TextRecBox is the mean box
score over ground-truth plates, so missed plates score zero and false positives
are not penalised. We additionally report mean IoU of matches, recall, the
fraction of matches with IoU $\ge0.5$, and the number of exactly (perfect,
$T=1$) and partially ($0<T<1$) read plates.
}

\section{Reference pipeline and baseline architectures}\label{sec:pipeline}
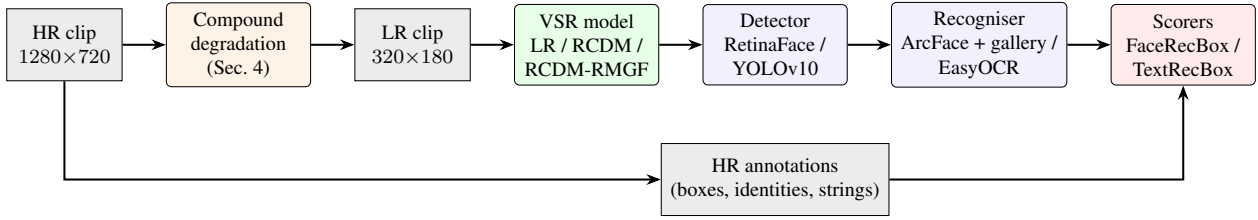
\begin{figure}[t]
\centering
\resizebox{\textwidth}{!}{%
\begin{tikzpicture}[
  node distance=6mm,
  blk/.style={draw,rounded corners=2pt,minimum height=10mm,minimum width=20mm,align=center,font=\scriptsize,fill=blue!6},
  io/.style={draw,minimum height=10mm,minimum width=16mm,align=center,font=\scriptsize,fill=gray!15},
  arr/.style={-{Stealth[length=2mm]},thick}]
\node[io] (hr) {HR clip\\$1280{\times}720$};
\node[blk,right=of hr,fill=orange!10] (deg) {Compound\\degradation\\(Sec.~\ref{sec:degr})};
\node[io,right=of deg] (lr) {LR clip\\$320{\times}180$};
\node[blk,right=of lr,fill=green!10] (vsr) {VSR model\\LR / RCDM /\\RCDM-RMGF};
\node[blk,right=of vsr] (det) {Detector\\RetinaFace /\\YOLOv10};
\node[blk,right=of det] (rec) {Recogniser\\ArcFace + gallery /\\EasyOCR};
\node[blk,right=of rec,fill=red!8] (sc) {Scorers\\FaceRecBox /\\TextRecBox};
\foreach \a/\b in {hr/deg,deg/lr,lr/vsr,vsr/det,det/rec,rec/sc} \draw[arr] (\a)--(\b);
\coordinate (mid) at ($(det.south)+(0,-1.1)$);
\node[io,below=8mm of det] (gt) {HR annotations\\(boxes, identities, strings)};
\draw[arr] (gt.east) -| (sc.south);
\draw[arr] (hr.south) |- (gt.west);
\end{tikzpicture}}
\caption{FANVIDv2 evaluation flow. The recognisers are fixed and used off the
shelf; only the VSR stage varies between systems. Ground truth is always
annotated on the HR clip.}\label{fig:flow}
\end{figure}

\textbf{Recognition pipeline.} Faces: LR (or super-resolved) clip
$\rightarrow$ RetinaFace detection $\rightarrow$ ArcFace embedding $\rightarrow$
cosine similarity to the gallery. Plates: clip $\rightarrow$ YOLOv10
\citep{yolov10} detection $\rightarrow$ EasyOCR \citep{easyocr} transcription.
Recognisers are used without fine-tuning (Fig.~\ref{fig:flow}).

\subsection{RCDM}
RCDM (Residual Convolutional Deformable Memory)
\citep{viswanathan2025lowresource} takes a group of $2k{+}1$ LR frames
$\{I^{\mathrm{LR}}_{t-k},\dots,I^{\mathrm{LR}}_{t+k}\}$ ($k{=}3$, a seven-frame
window) and emits one HR frame $\hat I^{\mathrm{HR}}_t$
(Figs.~\ref{fig:rcdm}--\ref{fig:rcdmdetail}). It combines three components.

\emph{(i) Wavelet conditioning.} A single-level 2D Haar discrete wavelet
transform of the central frame,
\begin{equation}
\mathrm{DWT}(x)=\{LL,LH,HL,HH\},
\end{equation}
gives a low-frequency approximation and horizontal, vertical and diagonal detail
sub-bands at half resolution. The sub-bands are upsampled to feature-map
resolution, concatenated with the features and, in every ConvNeXt fusion block
\citep{liu2022convnext}, linearly projected and added to the residual branch
(wavelet-aware fusion). Removing this conditioning lowers SSIM by roughly 0.012
on REDS4 and visibly softens edges. The wavelet branch follows the sparse,
multi-scale token mixing of WaveMix \citep{jeevan2023wavemix}.

\emph{(ii) Deformable alignment.} The $2k{+}1$ frames are concatenated
channel-wise ($3(2k{+}1)\times H\times W$) and passed through shallow 3D
convolutions. A modulated deformable convolution layer (DCNv2)
\citep{dai2017deformable,zhu2019dcnv2},
\begin{equation}
y(p_0)=\sum_{p_n\in\mathcal R}w(p_n)\,x(p_0+p_n+\Delta p_n)\,\Delta m_n,\quad\Delta m_n\in[0,1],
\end{equation}
fuses spatial detail and aligns neighbouring frames \emph{implicitly} through
learned offsets $\Delta p_n$, replacing explicit optical-flow estimation.

\emph{(iii) Residual memory tensor.} The aligned feature map $F_t$ updates a
recurrent memory through a learnable single-pole IIR rule,
\begin{equation}
M_t=\alpha M_{t-1}+(1-\alpha)F_t,\qquad\alpha\in(0,1)\ \text{learnable},
\end{equation}
which reinforces temporally stable structure and damps per-frame noise without
the gates of ConvLSTM/ConvGRU. $[F_t,M_t]$ passes through a stack of ConvNeXt
blocks, and a pixel-shuffle head (two $\times2$ stages for $\times4$ SR) adds
the result to a bicubic skip connection.

With $k{=}3$, RCDM has about 2.3\,M parameters and 281 GFLOPs per output frame
at $180\times320\rightarrow720\times1280$, roughly an order of magnitude below
VRT and RVRT. Under bicubic LR it reaches SSIM 0.9175 on REDS4. For paired
supervision the loss is
$\mathcal L=\lambda_1\mathcal L_{\mathrm{Char}}+\lambda_2(1-\text{MS-SSIM})+\lambda_3\mathcal L_{\mathrm{perc}}$
with a Charbonnier term ($\epsilon=10^{-3}$), MS-SSIM \citep{wang2003msssim}
and a VGG-19 \texttt{relu3\_3} perceptual term \citep{johnson2016perceptual},
$(\lambda_1,\lambda_2,\lambda_3)=(1,0.5,0.05)$.

\begin{figure}[t]
\centering
\includegraphics[width=\textwidth]{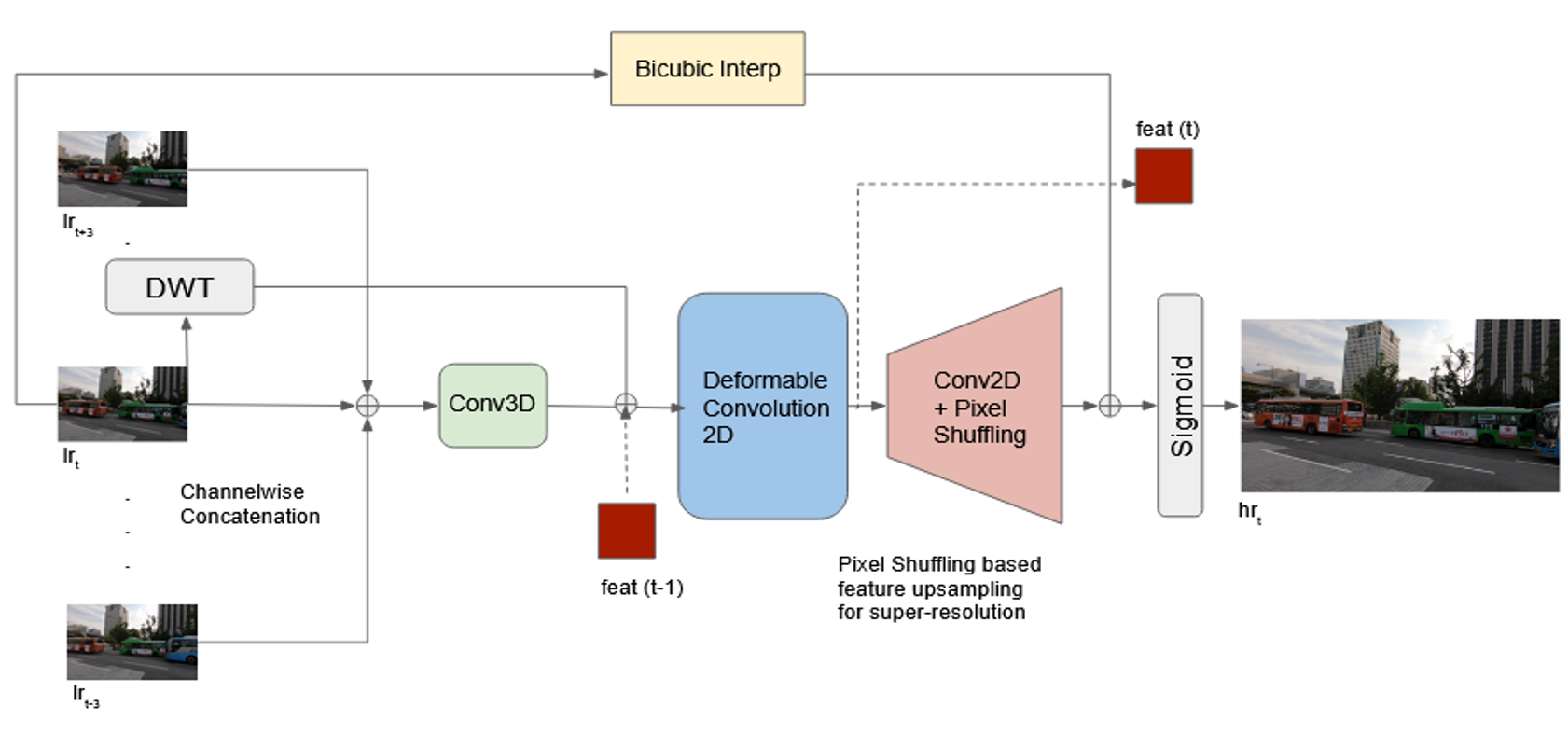}
\caption{RCDM base architecture: the frame window is concatenated and passed
through Conv3D; a 2D DWT of the central frame is added before a 2D deformable
convolution that jointly fuses and aligns; the recurrent memory
($\mathrm{feat}_{t-1}\rightarrow\mathrm{feat}_t$) feeds back into the fusion; a
Conv2D + pixel-shuffle head upsamples and a bicubic skip is added before the
output non-linearity.}\label{fig:rcdm}
\end{figure}

\begin{figure}[t]
\centering
\includegraphics[width=\textwidth]{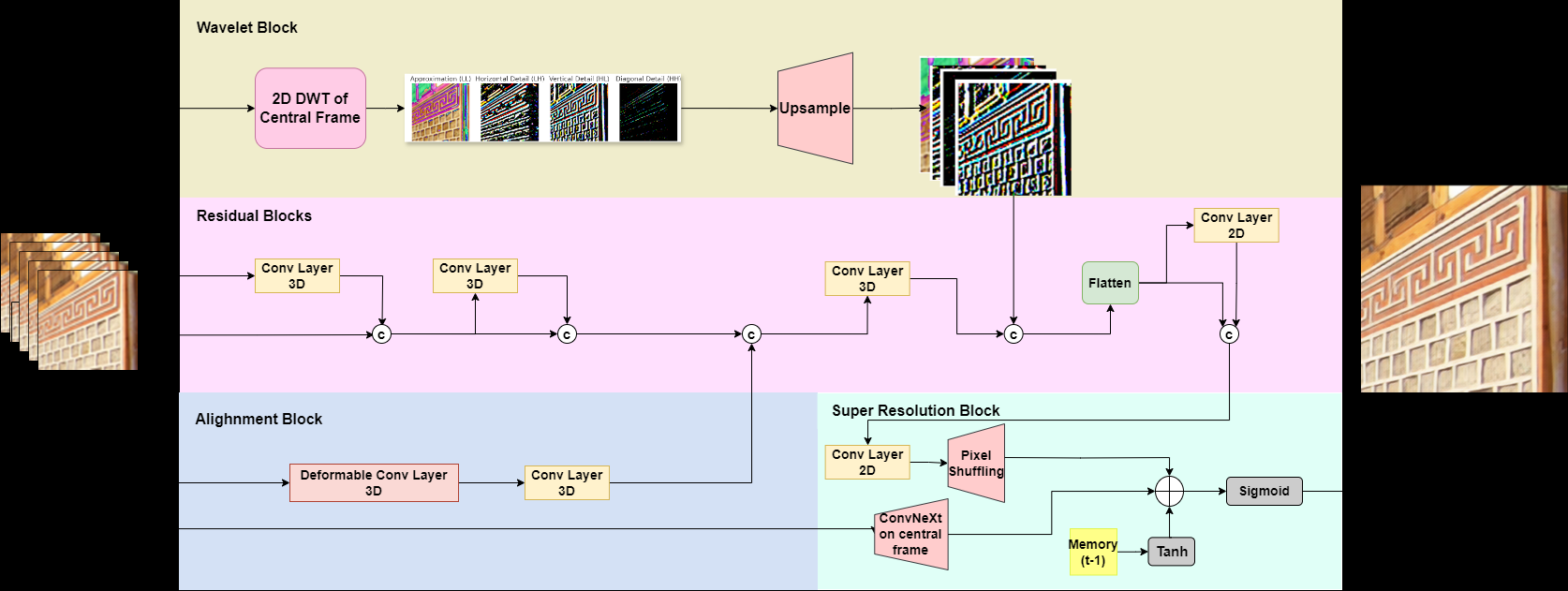}
\caption{Detailed RCDM pipeline. The wavelet block provides an upsampled
sub-band stream from the central frame; residual 3D convolution blocks and a 3D
deformable alignment block are concatenated (c); the super-resolution block
combines a pixel-shuffle path with a ConvNeXt path on the central frame and the
tanh-gated memory from $t{-}1$ before the final sigmoid.}\label{fig:rcdmdetail}
\end{figure}

\subsection{Residual-map gated fusion (RCDM-RMGF)}
Under heavy degradation the previous-frame state is unreliable where
consecutive frames differ strongly (large $|I_t-I_{t-1}|$), e.g.\ around moving
vehicles. RMGF adds a lightweight gating network (fewer than 0.05\,M
parameters; 2.35\,M in total) that maps the inter-frame residual to a gate on
the memory contribution, suppressing $M_{t-1}$ in such regions. On a held-out
split of REDS with the second-order degradation, RMGF improves validation
PSNR from 22.48 to 22.58\,dB and SSIM from 0.591 to 0.598 over pure RCDM.

\subsection{Modular family}
The RCDM backbone exposes configuration axes: propagation topology
(sliding-window, recurrent, or a 1.27\,M-parameter light-recurrent variant for
edge use); reconstruction head (pixel-shuffle or Laplacian-pyramid refinement
\citep{lai2017laplacian}, $+0.12$\,M parameters); block type (ConvNeXt, Swin or
WaveMix); and dataset adaptation (a 0.6\,M-parameter slim configuration).
The MRI work in the associated thesis reuses the WaveMix variant. For plates, a
text-specialised variant additionally uses glyph-focused patchification and an
OCR-coupled objective
$\mathcal L_{\mathrm{text}}=\lambda_c\mathcal L_{\mathrm{Char}}+\lambda_p\mathcal L_{\mathrm{perc}}+\lambda_s(1-\mathrm{SSIM})+\lambda_o\mathcal L_{\mathrm{OCR}}$,
with a transformer OCR recogniser \citep{li2023trocr,du2020ppocr} at inference;
it is not part of the results below, which use the fixed EasyOCR recogniser for
all systems.

\subsection{Baselines}
(i)~\emph{LR}: recognisers are applied to the LR clip upsampled bicubically.
(ii)~\emph{RCDM}: the 2.3\,M-parameter model above, trained on REDS with the
second-order degradation of Section~\ref{sec:degr} and applied to FANVIDv2
without further adaptation. (iii)~\emph{RCDM-RMGF}: RCDM with residual-map
gated fusion, trained the same way.

\section{Results}\label{sec:results}
\begin{table}[t]
\caption{Face task on the FANVIDv2 test split (hard-IoU scorer,
$139{,}093$ ground-truth face boxes).}\label{tab:face}
\centering\small
\begin{tabular}{@{}lcc@{}}
\toprule
Metric & LR & RCDM\\
\midrule
FaceRecBox & 0.6864 & \textbf{0.7222}\\
Precision & 0.9077 & 0.9133\\
Recall & 0.8872 & 0.9019\\
F1 & 0.8973 & 0.9076\\
Mean IoU of matches & 0.9363 & 0.9685\\
Identity accuracy on matches & 0.8435 & 0.8693\\
Correct identity matches & 104,094 & 109,053\\
\bottomrule
\end{tabular}
\end{table}

\begin{table}[t]
\caption{Plate task on the FANVIDv2 test split (aligned scorer). All conditions
use the same HR annotations as ground truth. The LR and SR outputs are evaluated
on different numbers of common frames (14,189 vs.\ 15,758), so absolute counts are
not directly comparable; rates per ground-truth box are. $^\ddagger$ RMGF counts
were produced by the same scorer; its ground-truth box count is not tabulated
here.}\label{tab:plate}
\centering\small
\setlength{\tabcolsep}{4pt}
\begin{tabular}{@{}lccc@{}}
\toprule
Metric & LR & RCDM & RCDM-RMGF\\
\midrule
TextRecBox & 0.3088 & 0.3667 & \textbf{0.3801}\\
Mean IoU of matches & 0.6067 & 0.6958 & \textbf{0.7113}\\
Recall & 0.9871 & 0.9931 & ---\\
IoU $\ge0.5$ (\%) & 74.69 & 77.61 & ---\\
IoU $\ge0.3$ (\%) & 76.12 & 78.84 & ---\\
Ground-truth boxes & 31,739 & 34,996 & ---\\
Predicted boxes & 31,331 & 34,763 & ---\\
Perfect reads (count) & 715 & 985 & \textbf{1,002}$^\ddagger$\\
Perfect reads per 1000 GT boxes & 22.5 & 28.1 & ---\\
Partial reads (count) & 7,334 & 9,031 & ---\\
\bottomrule
\end{tabular}
\end{table}

\textbf{Faces.} Super-resolution raises FaceRecBox from 0.6864 to 0.7222, an
absolute gain of 3.6 points (Table~\ref{tab:face}). Both precision and recall
increase, so the gain is not a trade of one error type for another. Matched
detections become better localised (mean IoU 0.936 to 0.969), identity
accuracy on matched faces rises from 84.35\% to 86.93\%, and the number of
correct identity matches grows by 4,959. Fig.~\ref{fig:qual}(a,b) shows a face
that is not detected in the LR frame and is detected after VSR.

\textbf{Plates.} TextRecBox rises from 0.3088 to 0.3667 with RCDM (a relative
gain of 18.7\%) and to 0.3801 with RCDM-RMGF (23.1\%) (Table~\ref{tab:plate}).
The count of exactly read plates grows from 715 to 985 (+37.8\%) and to 1,002.
Because SR output is evaluated on more common frames than LR output, the count
comparison is confounded; normalised per ground-truth box, exact reads rise
from 22.5 to 28.1 per 1000 boxes (+24.9\%). Mean localisation IoU rises from
0.607 to 0.696 with RCDM and 0.711 with RCDM-RMGF. Recall is almost unchanged
(0.987 to 0.993) because the detector already finds most plates in LR
frames; the gains come mainly from reading them correctly and localising them
more tightly. Fig.~\ref{fig:qual}(c,d) shows a plate that is detected but
unreadable in the LR frame and read after VSR.

\textbf{Fidelity and recognition.} On the plate clips, RCDM improves per-frame
fidelity over LR input by $+1.71$\,dB PSNR ($24.26\rightarrow25.97$), $+0.074$
SSIM ($0.733\rightarrow0.807$) and MicroSSIM \citep{ashesh2024microssim}
($0.672\rightarrow0.767$), and reduces LPIPS \citep{zhang2018lpips} from 0.437
to 0.137. Fidelity and recognition therefore move together here, but they are
not equivalent: in our model comparisons PSNR and SSIM correlate only weakly with
TextRecBox (Spearman $\rho\approx0.3$) and moderately with FaceRecBox
($\rho\approx0.6$), which is the motivation for scoring by recognition. The
script \texttt{code/compute\_fidelity\_and\_correlation.py} recomputes these
statistics from per-frame results.

\begin{figure}[t]
\centering
\begin{tabular}{@{}cc@{}}
\includegraphics[width=0.47\textwidth]{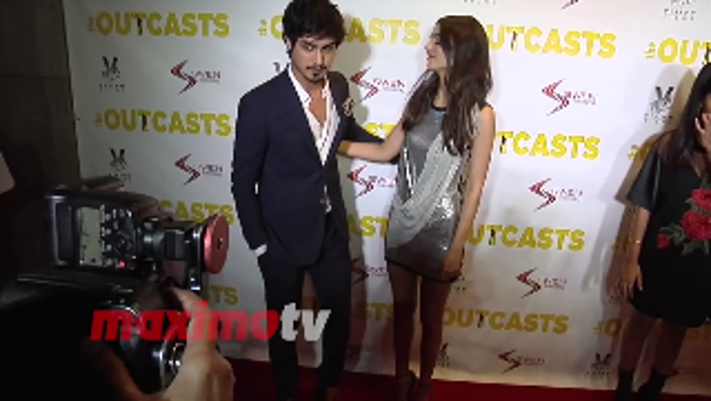} &
\includegraphics[width=0.47\textwidth]{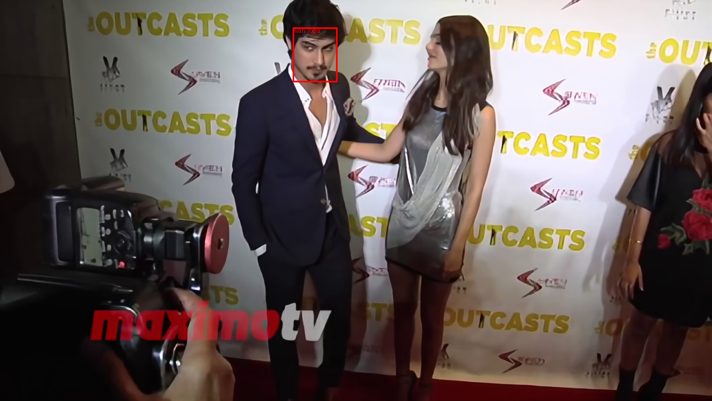}\\
\small (a) LR: face not detected & \small (b) RCDM: face detected\\[3pt]
\includegraphics[width=0.47\textwidth]{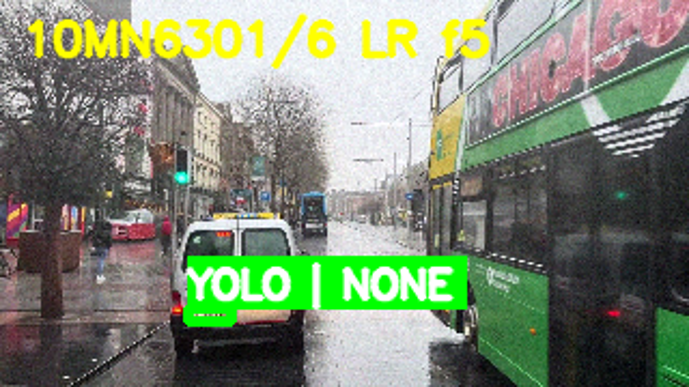} &
\includegraphics[width=0.47\textwidth]{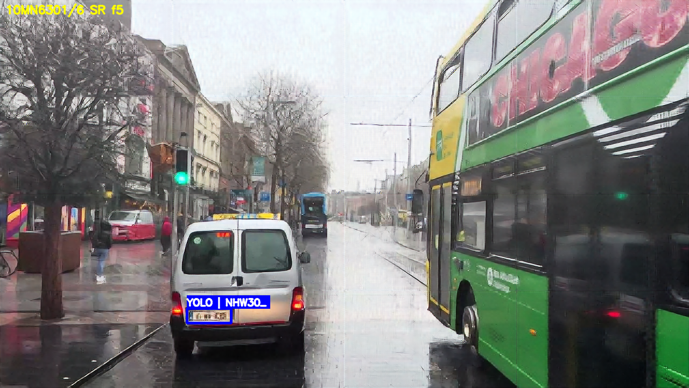}\\
\small (c) LR: plate detected, OCR fails & \small (d) RCDM: plate read\\
\end{tabular}
\caption{Examples from the FANVIDv2 test split.}\label{fig:qual}
\end{figure}

\section{Discussion}
\textbf{What the benchmark measures.} FANVIDv2 asks whether the output of a
VSR model can be recognised, not whether it looks like the original. The two
tasks stress different properties: on faces the gain comes with better
localisation and identity matching, on plates with correct reading of characters
that the detector had already located.

\textbf{Limitations.} FANVIDv2 is an evaluation benchmark, not a large-scale
recognition dataset: 48 identities and 360 plate strings. The face subset is
drawn from celebrity footage, chosen for legal clearance and gallery-image
availability, which over-represents frontal pose and good illumination relative
to operational settings. The degradation is synthetic; a camera-native LR test
set would strengthen external validity. Plate strings are Latin alphanumeric
only. The recognisers are used off the shelf, so the results measure what VSR
contributes to a fixed pipeline, not the best achievable recognition. Two plate
strings occur in both plate splits. LR and SR plate outputs are evaluated on
different numbers of common frames (Table~\ref{tab:plate}), so we report
normalised rates. Only three systems are evaluated here; results for stronger
VSR baselines and transformer OCR recognisers should be produced with the
released scorers.

\textbf{Ethics.} All subjects are public figures in publicly available
footage. No video is redistributed; frames are regenerated from public URLs
and LR frames are $320\times180$. We discourage operational surveillance or
law-enforcement use without ethical review and compliance with local privacy
law.

\section{Conclusion}
FANVIDv2 evaluates video super-resolution by face identification and plate
reading under randomised compound degradation. A lightweight, 2.3\,M-parameter
VSR baseline improves both tasks (FaceRecBox $0.686\rightarrow0.722$;
TextRecBox $0.309\rightarrow0.367$), with a further plate gain from
residual-map gated fusion. We release annotations, metadata, degradation and
download scripts and scorers so that VSR methods can be compared by what they
make recognisable.

\section*{Data and code availability}\label{sec:release}
Annotations, gallery metadata and download scripts are hosted in the FANVID
repository: \url{https://huggingface.co/datasets/kv1388/FANVID-Face_and_License_Plate_Recognition_in_Low-Resolution_Videos}.
The degradation generators, the two scorers and the fidelity/correlation
script are provided in the \texttt{code/} directory accompanying this
submission.

\section*{Declaration of generative AI and AI-assisted technologies}
Generative AI tools were used to assist with language editing, consistency
checks and manuscript preparation. The authors reviewed and edited the
manuscript and are responsible for its final content.

\bibliographystyle{plainnat}
\bibliography{refs}
\end{document}